\documentclass[10pt,twocolumn,letterpaper]{article}
\pdfoutput=1

\usepackage{color}

\usepackage{url}

\definecolor{noah_color}{rgb}{0.2,.64,0}
\definecolor{kb_color}{rgb}{1,.14,0}
\definecolor{sean_color}{rgb}{1,.5,0}
\definecolor{paul_color}{rgb}{.6,.25,.9}
\definecolor{todo_color}{rgb}{.7,.7,.1}
\definecolor{new_color}{rgb}{.1,.7,.8}

\iftrue
  \newcommand{\noah}[1]{\textsf{\textcolor{noah_color}{[{\bf Noah says}: #1]}}}
  \newcommand{\kb}[1]{\textsf{\textcolor{kb_color}{[{\bf KB says}: #1]}}}
  \newcommand{\sean}[1]{\textsf{\textcolor{sean_color}{[{\bf Sean says}: #1]}}}
  \newcommand{\paul}[1]{\textsf{\textcolor{paul_color}{[{\bf Paul says}: #1]}}}
  \newcommand{\todo}[1]{{\sf\textcolor{todo_color}{TODO: #1}}}
  
  \newcommand{\kbrem}[1]{{\textcolor{kb_color}{\em{KB: #1}}}}
  \newcommand{\paulrem}[1]{{\textcolor{paul_color}{\em{Paul: #1}}}}
  \newcommand{\noahrem}[1]{{\textcolor{noah_color}{\em{Noah: #1}}}}
\else
  \newcommand{\noah}[1]{}
  \newcommand{\kb}[1]{}
  \newcommand{\sean}[1]{}
  \newcommand{\paul}[1]{}
  \newcommand{\todo}[1]{}
  \newcommand{\kbrem}[1]{}
  \newcommand{\paulrem}[1]{}
  \newcommand{\noahrem}[1]{}
\fi

\newcommand{\leaveout}[1]{}

\newcommand{\etal}{{et~al.}}

\newcommand{\tablesize}[1]{{\footnotesize{#1}}}

\newcommand{\kw}{\text{M2}\xspace}

\newcommand{\Reals}{\mathbb{R}}

\usepackage{amsopn}

\usepackage{cvpr}
\usepackage{times}
\usepackage{graphicx}
\usepackage{epstopdf}
\usepackage{amsmath}
\usepackage{amssymb}

\usepackage{float}
\usepackage{microtype}
\usepackage{multirow}
\usepackage{booktabs}
\usepackage{nth}
\usepackage{array}
\usepackage{xcolor}
\usepackage{calligra}
\usepackage{rotating}
\usepackage{aurical}
\usepackage{pbsi}
\usepackage{yfonts}
\usepackage[T1]{fontenc}
\newfont{\hge}{hge scaled 1500}
\usepackage{titling}

\usepackage[pagebackref=true,breaklinks=true,letterpaper=true,colorlinks,bookmarks=false]{hyperref}

\cvprfinalcopy 

\def\cvprPaperID{1774} 

\begin{document}

\title{From {\bsifamily A} to {\hge Z}: Supervised Transfer of Style and Content Using Deep Neural Network Generators}

\author{%
Paul Upchurch
\qquad
Noah Snavely
\qquad
Kavita Bala
\\
Department of Computer Science, Cornell University
\\
{\tt\small \{paulu,snavely,kb\}@cs.cornell.edu}
}

\maketitle

\begin{abstract}

We propose a new neural network architecture for solving single-image
analogies---the generation of an entire set of stylistically similar
images from just a single input image. Solving this problem requires
separating image style from content. Our network is a modified
variational autoencoder (VAE) that supports supervised training of
single-image analogies and in-network evaluation of outputs with a
structured similarity objective that captures pixel covariances.  On
the challenging task of generating a 62-letter font from a single
example letter we produce images with 22.4\% lower dissimilarity to
the ground truth than state-of-the-art.

\end{abstract}

\vspace{-4pt}
\section{Introduction}

Separating image style from content is an important problem in vision
and graphics. One key application is {\em analogies}: if we can
separate an image into style and content factors, we can generate a
new image by holding style fixed and modifying the content (resulting
in an image with different semantics but the same style), or vice
versa.  For instance, given a letter `{\bsifamily A}' the goal would
be to generate 
all of the letters `{\bsifamily B}', `{\bsifamily C}', ... in the same style.
Or, given a stylized image (e.g., Instagram filtered), the goal could
be to recover the unfiltered image or switch to a different filter.

We call this kind of analogy a {\em single-image analogy} since only
a single image is given as input. Clearly, producing `{\bsifamily B}'
given only `{\bsifamily A}' is impossible without drawing on some prior
knowledge, so learning a prior over either style or content will
play a role in solving single-image analogies.

Kingma et al~\cite{kingma2014semi} observed that deep neural networks
can learn latent spaces in which reasoning about single-image analogies
is possible.
Style is implicitly preserved in the organization of
the learned embedding space. They produce compelling results, yet they
did not directly optimize for preserving style or producing analogies,
nor do they quantitatively evaluate the analogies they produce.

\begin{figure}[H]
\begin{center}
  \includegraphics[width=0.95\linewidth]{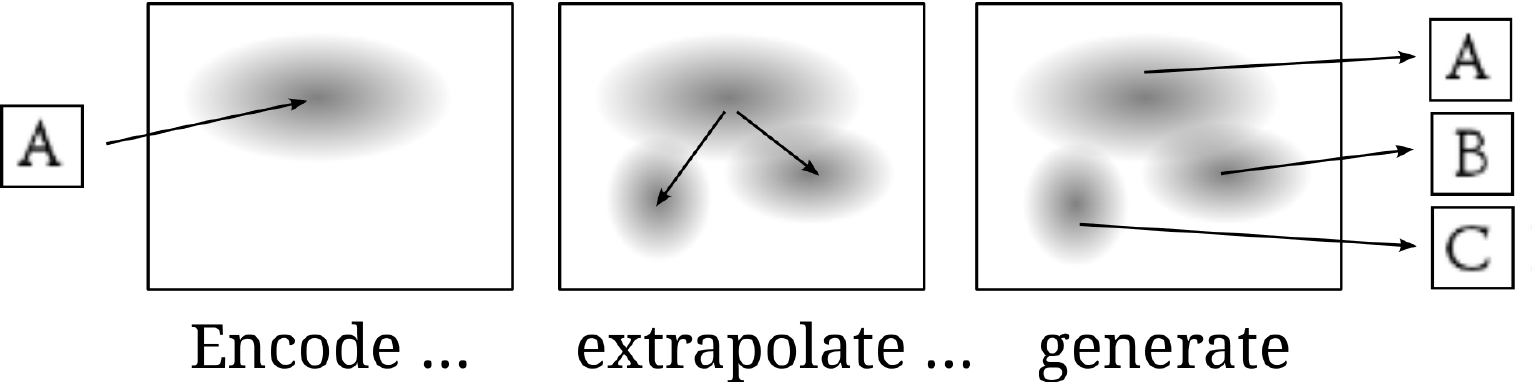}
\end{center}
\vspace{-6pt}
\caption{%
  \textbf{Analogical reasoning.} This figure illustrates how our method produces a font from a single letter. First, the input image (e.g., `A') is encoded to a Gaussian distribution in a latent space. Next we extrapolate the latent distribution to new latent distributions which describe other letters of the same font. Finally, we generate new images by sampling each latent distribution and mapping the sampled point to an image.}
\label{fig:teaser}
\end{figure}

We extend the work of~\cite{kingma2014semi} by introducing a
new deep neural network architecture for better single-image analogies.
Our key insight is that having training data grouped into {\em style
sets} (a sequence of images with known content and consistent style)
allows direct optimization of a style embedding space and reasoning
about single-image analogies.

We demonstrate our method on a new, large-scale dataset of 1,839 fonts. We
experiment on 10-class digits and 62-class alphanumerics and show that
our method consistently outperforms previous work under a variety of
experimental settings. Though our demonstration application is font
generation, our architecture is not specific to any domain.

Our insight is to train on supervised style sets and optimize for
image quality. Thus, our contributions are:
\begin{itemize}
\setlength\itemsep{-4pt}
\item A neural network model~\cite{kingma2014semi}:
Our model has a latent distribution extrapolation
layer and two adversarial
networks~\cite{goodfellow2014generative} (class and imposter
discriminators).
\item Improved image quality with SSIM~\cite{wang2004image}: Our model optimizes generated images for structured similarity (i.e., pairwise pixel covariances) to ground truth.
\item A challenging image analogy dataset: Our dataset has more classes, more styles and more subtle style cues compared to MNIST~\cite{lecun1998gradient} and Street View House Numbers (SVHN)~\cite{netzer2011reading}.

\end{itemize}

\section{Related Work}

Image analogies are part of a larger research area of image generation
which has attracted much attention. Image analogies support a variety
of applications (e.g., super-resolution, texture transfer,
texture synthesis and automatic artistic filtering). The generalized image
analogy problem can be phrased as ``A is to B as C is to D''. The goal
is to produce image D.

We first describe works which require A, B and C as image inputs.
Hertzmann et al~\cite{hertzmann2001image} proposed a method where they
suppose the pixelwise content of (A, B) and (C, D) pairs are nearly
identical. Thus, they dynamically construct a high-quality D by finding
pixelwise matches and transferring style.  Under similar constraints,
Memisevic and Hinton~\cite{memisevic2007unsupervised} proposed an
unsupervised model which learns the image transformations.  Taylor
et al~\cite{taylor2010convolutional} present a third-order restricted
Boltzmann machine (RBM) which can generate analogs without requiring
(A, B) and (C, D) pairs to be pixelwise dependent.
 Although not intended
for images, the \texttt{word2vec} project~\cite{mikolov2013distributed}
found that learned latent spaces can easily solve analogical reasoning
tasks. Sadeghi et al~\cite{sadeghi2015visalogy} do not generate images but they
learn a latent space for solving analogy queries.

A parallel work\footnote{Our work is a 2015 November 6 submission to CVPR 2016.} by Reed et al~\cite{reed2015deep} develop a deep neural network for visual analogies with three input images. They propose three forms of extrapolation and a disentangled feature representation.

Recently, Gatys et al~\cite{gatys2015neural} have proposed a
scalable, high-quality method which uses the layered features of
a deep convolutional neural network (CNN) to
separate content from style. They do not require
image A. Given only images B and C, they produce an image D which matches
the content of C and the style of B. They draw upon knowledge learned by
training a CNN on over 1 million images.

Single-image analogies are possible. To our knowledge, only Kingma
et al~\cite{kingma2014semi} claim to produce style and content factored image analogies from
a single image. Given only image B they produce digits
which match the style of B while adhering to a specified content class
(restricted to content classes which have been seen during training).
Tenenbaum and Freeman~\cite{tenenbaum2000separating} propose an
extrapolation method that produces letters of a font given only examples of
other letters of the same font. However, they only demonstrate results
when 53 input letters are provided.

Kulkarni et al~\cite{kulkarni2015deep} present an inverse graphics model that disentangles pose
and lighting. They propose a novel training procedure for factoring their latent space.

Using SSIM as an optimization target is not a new
idea~\cite{wang2004stimulus, channappayya2006linear, oztireli2015perceptually}.
Indeed, much work as been done on analysis of the quasi-convexity
of SSIM~\cite{channappayya2008design, brunet2012mathematical}.
Regarding generative models, Ranzato and Hinton~\cite{ranzato2010modeling}
first argued that capturing dependencies between pixels
is important. Subsequently, there have been models which aim to capture pairwise dependencies in natural images~\cite{Ranzato2010, ICML2011Courville_591, hao2012gated}.

\begin{figure}[tb]
\begin{center}
  \includegraphics[width=0.95\linewidth]{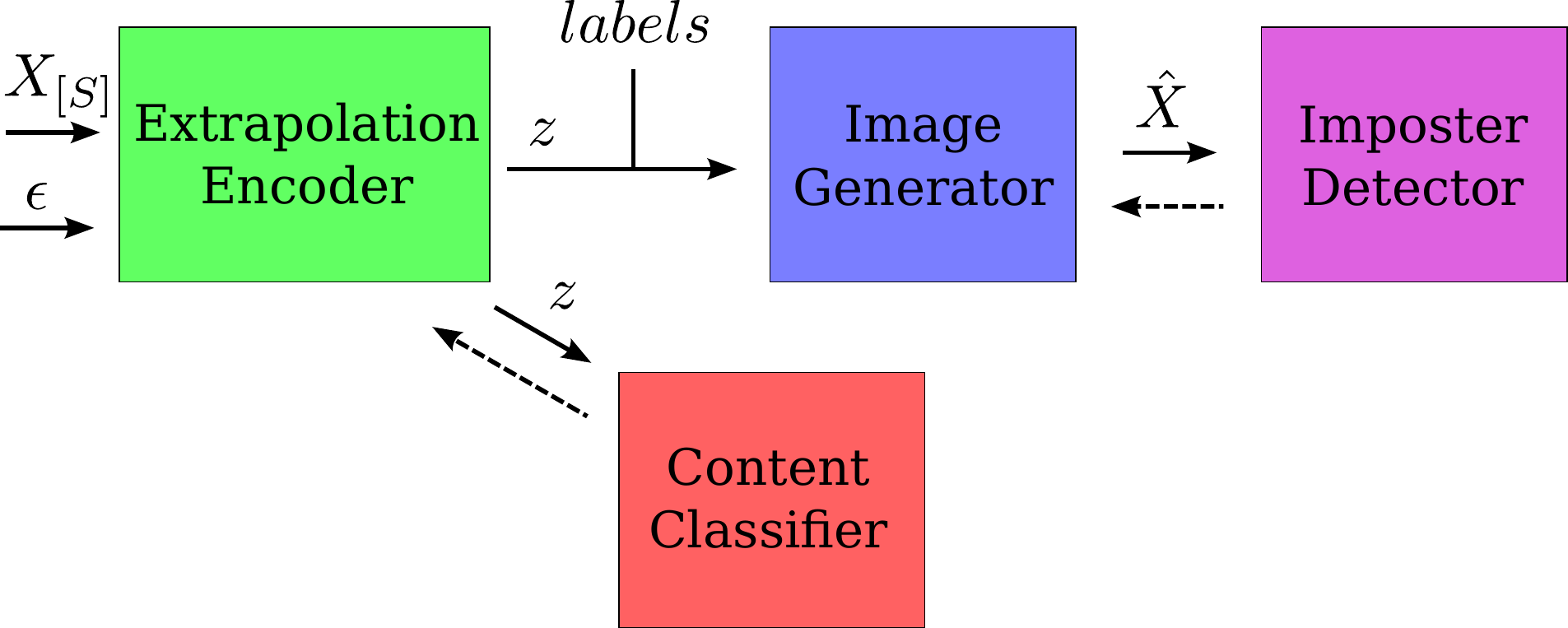}
\end{center}
\vspace{-6pt}
\caption{%
  \textbf{Model overview.} The variational encoder (green) takes a subset of
the style set images $X_{[S]}$ (example shown) and a noise variable $\epsilon
\sim \mathcal{N}(0,I)$ as input, and produces code $z$ for all members of the
style set, and combines the codes with identity class labels.  The generator
(blue) to image reconstructions $\hat{X}$ (example shown) for the entire style
set. An imposter detector adversary (purple) is used to improve the
reconstruction quality. An identity classifier adversary (red) is
used to promote class invariance on $z$ codes. Dotted lines indicate zones
which update weights with a negated gradient wrt loss functions in the
adversarial zones.
}
\label{fig:overview}
\end{figure}

\section{Overview}
\label{sec:method}

At its core, our method is an encoder / decoder pair (Figure~\ref{fig:teaser}). The encoder takes an image
and maps it to a point in a $Z$-dimensional embedding space $\Reals^Z$
(aka a {\bf latent space}).
Our method enforces structure on this space: it represents only style, but not content.
The decoder combines style information
from the embedding space with a content switch variable.
For example, if our domain is fonts, and we feed in `{\tt A}' to the encoder, we
get a point in the embedding space $\Reals^Z$ representing the style of `{\tt A}'.
If we pass this point to the decoder, but with the content switch `B', then the
decoder should produce `{\tt B}'---the letter `B', but in the style of `{\tt A}'.

We now describe our key contributions: a deep neural network
architecture, in-network and out-of-network image quality assessment
measures, and a new font dataset with high-quality ground truth for
analogical reasoning.

\medskip
\noindent{\bf Network architecture.}
Our neural network is a modified variational autoencoder
(VAE)~\cite{kingma2013auto, kingma2014semi}, a state-of-the-art
semi-supervised image generator. We introduce
VAEs
here and briefly review
their mathematics
in Section~\ref{sec:variationalencoder}.

A VAE encoder is trained to learn a mapping from images
to {\bf latent codes}.
These codes are not points, but rather multivariate Gaussians.
Mapping images to distributions conditions the latent space~\cite{kingma2013auto}.
The encoder must learn a mapping where
differing images rarely overlap. This yields a better organized latent space.
We then sample from a latent code to get a latent point that we can feed to the decoder.

Kingma \etal~\cite{kingma2014semi} demonstrated a {\bf conditional}
decoder model that can produce image analogies for MNIST and SVHN
digits.
Their decoder takes
both a latent point and a content variable (i.e., `1', `2', `3', etc.),
and is trained to produce styled images of the specified content.
This structure naturally lends itself to producing complete sets of digits,
given a single example.

Our idea is to use supervision to learn analogies. Since (in the case of fonts)
the content labels (`A', `B', ...) are available for all training examples,
we want to use those labels to learn to generate an entire style set (given some
subset as examples).
Our modifications to the standard VAE model are natural consequences of this key idea. As we describe
each change in the following paragraphs, we refer the reader to the colored zones in
Figure~\ref{fig:overview}, which shows a high-level overview of the
network. Also, components of our model are described mathematically in
Section~\ref{sec:model}.

\smallskip

\emph{Extrapolation Layer.}
Our input is only a subset of the style set so the latent codes for the
entire subset must be extrapolated.
In our model we extrapolate by taking linear combinations of the latent means and variances of the subset images to produce a latent mean and variance for each member of the style set.
This gives our model more expressive power than~\cite{kingma2014semi} which holds the latent mean and variance constant across all members. We call our method an extrapolation layer and find that gives a significant increase in performance.

\emph{SSIM Cost Function.}
The VAE conditional decoder (Figure~\ref{fig:overview} blue zone)
generates a complete style set by combining latent codes (sampled from
the latent distributions) with class labels.
Since pairwise pixel interactions are important in human perception and modeling
images~\cite{ranzato2010modeling}, our idea is to use an objective function based on
SSIM~\cite{wang2009mean}, an image quality measure that captures
pairwise dependencies.  We describe our objective function in
Section~\ref{sec:ssim} and compare against an $L_2$ objective
in Section~\ref{sec:psnrssim}.

\emph{Adversarial Sub-Networks}.
Ideally, the latent space should be a style embedding.
Image content should come from the content variable we pass to the decoder ---
not from the latent code.
We factor
content out of
the latent space with an adversarial~\cite{goodfellow2014generative}
content classifier connected to the encoder by a gradient reversal
layer~\cite{ganin2015unsupervised} (Figure~\ref{fig:overview} red
zone).
Gradient reversal layers
render a latent space invariant to a domain during training.
We explain how this
works in Section~\ref{sec:cls} and we find it gives a small performance
improvement (Section~\ref{sec:experimentadversary}).
We also experimented with adding an adversarial imposter detector to improve the visual quality
of the generated output (Figure~\ref{fig:overview} purple zone,
Section~\ref{sec:img}).

\medskip
\noindent{\bf A new dataset for quantitative evaluation.}
We require training data with both style and content annotations.
Fortunately, human designers have deliberately
created a multitude of complete sets of stylistically consistent fonts.
Furthermore, these have subtle, challenging style variations. It may seem overly
challenging to generate an entire font from a single letter, but this has
been demonstrated by~\cite{campbell2014learning} (although they use
domain-specific knowledge).
Compared to MNIST and SVHN, our dataset has more classes,
and emphasizes subtle style cues such as curvature, stroke, width, slant, and serifs. We
describe our dataset further in Section~\ref{sec:dataset}.

\begin{figure*}[tb]
\begin{center}
  \includegraphics[width=0.95\linewidth]{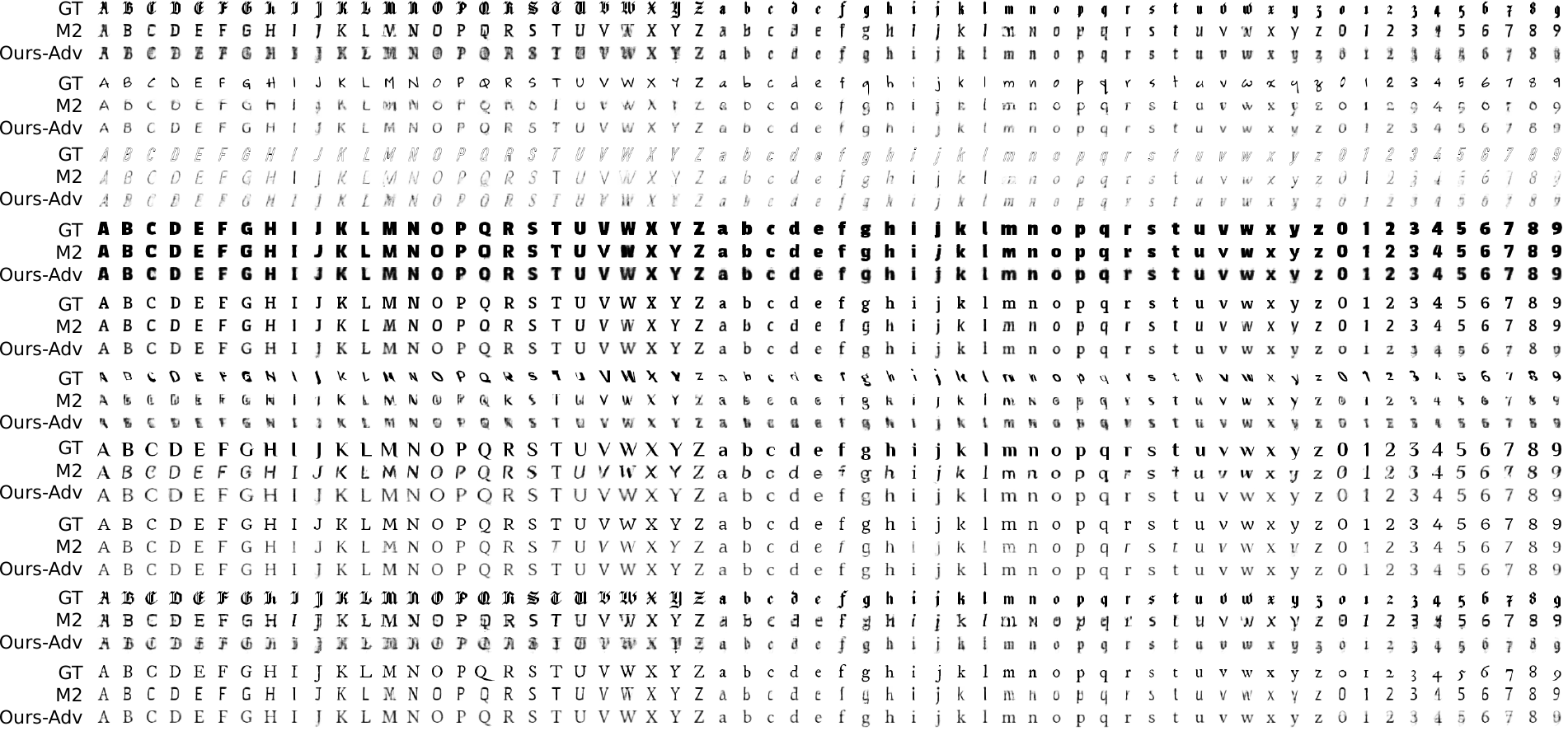}
\end{center}
\vspace{-6pt}
\caption{%
  \textbf{Random visual comparisons.} We visualize ground truth (GT), VAE (\kw)
  and our full model (Ours-Adv) on randomly selected test fonts. The input is `A'. Our method captures the blackletter style better (first, ninth font). The backwards-slanted sixth font poses problems for both methods. \kw cannot sustain the style of the `A' on the seventh font.
}
\label{fig:randomeval}
\end{figure*}

\begin{figure}[tb]
\begin{center}
  \includegraphics[width=0.95\linewidth]{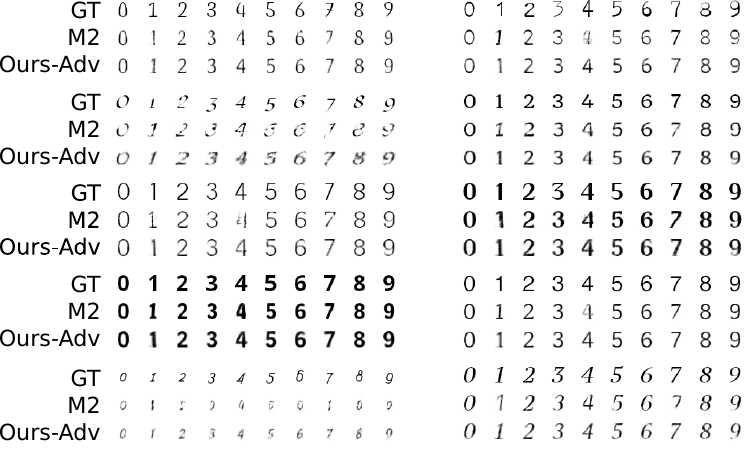}
\end{center}
\vspace{-6pt}
\caption{%
  \textbf{Random visual comparisons.} We visualize ground truth (GT), VAE (\kw)
  and our full model (Ours-Adv) on randomly selected test fonts. On many fonts our `4' and `7' are noticeably better. The small font (bottom-left) seems to be challenging for both methods. The thin parts of stokes seem to disappear for \kw (bottom-right font).
}
\label{fig:randomevaldigit}
\end{figure}

\medskip
\noindent{\bf Evaluation.}
We give
extensive quantitative comparisons of our method to variational
autoencoders
in
Section~\ref{sec:experiments}.
Our method outperforms
standard variational autoencoders
under
many
conditions. For qualitative
evaluation, we offer many analogical reasoning examples in
Figures~\ref{fig:psnrssim},~\ref{fig:digits},~\ref{fig:randomeval},
and~\ref{fig:randomevaldigit}.

We also perform experiments to justify our
modifications
to the VAE
model. We investigate the changes in performance when the
extrapolation layer and adversarial discriminators are removed from
the network. Those experiments are described in
Section~\ref{sec:noearly} and~\ref{sec:experimentadversary}.

Finally, we do not use domain-specific knowledge, so our model is extensible to other domains.

\section{Font Dataset}
\label{sec:dataset}

\begin{figure}[tb]
\begin{center}
  \includegraphics[trim={0 0 1840px 0},clip,width=0.95\linewidth]{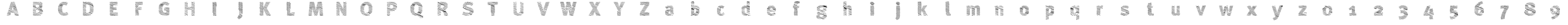}
  \includegraphics[trim={0 0 1840px 0},clip,width=0.95\linewidth]{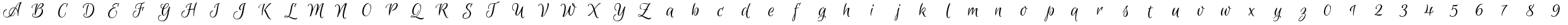}
  \includegraphics[trim={0 0 1840px 0},clip,width=0.95\linewidth]{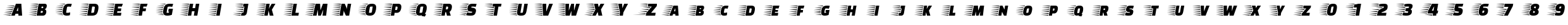}
  \includegraphics[trim={0 0 1840px 0},clip,width=0.95\linewidth]{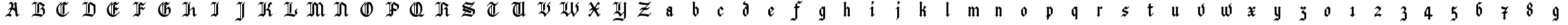}
\end{center}
\begin{center}
  \includegraphics[trim={0 0 1440px 0},clip,width=0.95\linewidth]{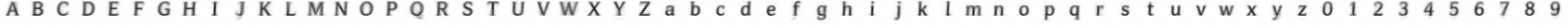}
  \includegraphics[trim={1040px 0 400px 0},clip,width=0.95\linewidth]{figures/dataset/visualizeaveragefont.png}
  \includegraphics[trim={2080px 0 -640px 0},clip,width=0.95\linewidth]{figures/dataset/visualizeaveragefont.png}
\end{center}
\vspace{-6pt}
\caption{%
  \textbf{Font dataset.} Top four rows: subsets of selected decorative, script and blackletter fonts to illustrate the variety of styles present in the dataset. Bottom: the average font, which looks normal since the majority of fonts are serif or sans serif.}
\label{fig:dataset}
\end{figure}

We have created a dataset of 1,839 fonts to provide quality ground
truth for analogical reasoning.
We used fonts from \texttt{openfontlibrary.org}
as well as the fonts used by~\cite{o2014exploratory}. Our dataset
includes serif, sans serif, blackletter, calligraphic, script and
decorative font styles. From each font we produce a style set of 62
images---26 uppercase letters, 26 lowercase letters and 10 digits.


We ensure each font is unique by eliminating any duplicate image sets.
However, duplicate letters may still exist, which would add unwanted bias to our dataset.
We address this by clustering letters:
We merge two clusters if at least one pair of letters (1) are identical or (2) come
from the same file (professional font creators often store
related fonts together) or (3) have the same font family (based on
metadata in the font file). We selected a random
subset of clusters such that 191 fonts were in
the test set. Of the remainder we use 92 fonts for validation and
1,556 for training. There are 102,176
training samples and 11,842 testing samples.

Our dataset is of a comparable size to MNIST, but contains 6.2 times as many
classes, as well as some unusual styles. The dataset average is similar to
a typical sans serif font.
See Figure~\ref{fig:dataset} for selected examples of the
different styles.  Note that the decorative fonts in the first and
third rows would be difficult to generate with the method
of~\cite{campbell2014learning}, which relies on outline alignment.

On our dataset, successful analogical reasoning is well-defined. For
example, if given the image of letter `A' from Arial and asked to
produce a `B', then the unique correct answer is the Arial image of letter
`B'.

\section{Neural Network Model}
\label{sec:model}

In this section we mathematically define our problem and describe the components of our model. Let $X$
be a complete, ground truth style set of $M$ images. Let $X_{[S]}$ be
a specific subset of $X$ (e.g., if we select $S=\{\text{`A'}\}$ then
the given letter is always letter `A').  Given only $X_{[S]}$ we will
generate $\hat{X}$, a reconstruction of the entire style set. The goal
is to minimize the dissimilarity of $\hat{X}$ and $X$.

Our model can be divided into four zones as depicted in
Figure~\ref{fig:overview}.  The purpose of each zone is described in
Section~\ref{sec:method}. The green zone is a variational encoder and
extrapolation layer that produces latent codes given images.  The blue
zone is an image generator (aka a decoder) that produces an image when
given a latent code and a class label.  The purple zone is a
discriminator trained to detect image imposters. The red zone is an
$M$-way classifier trained to detect which class produced a given
latent code.

\begin{figure}[tb]
\begin{center}
  \includegraphics[width=0.95\linewidth]{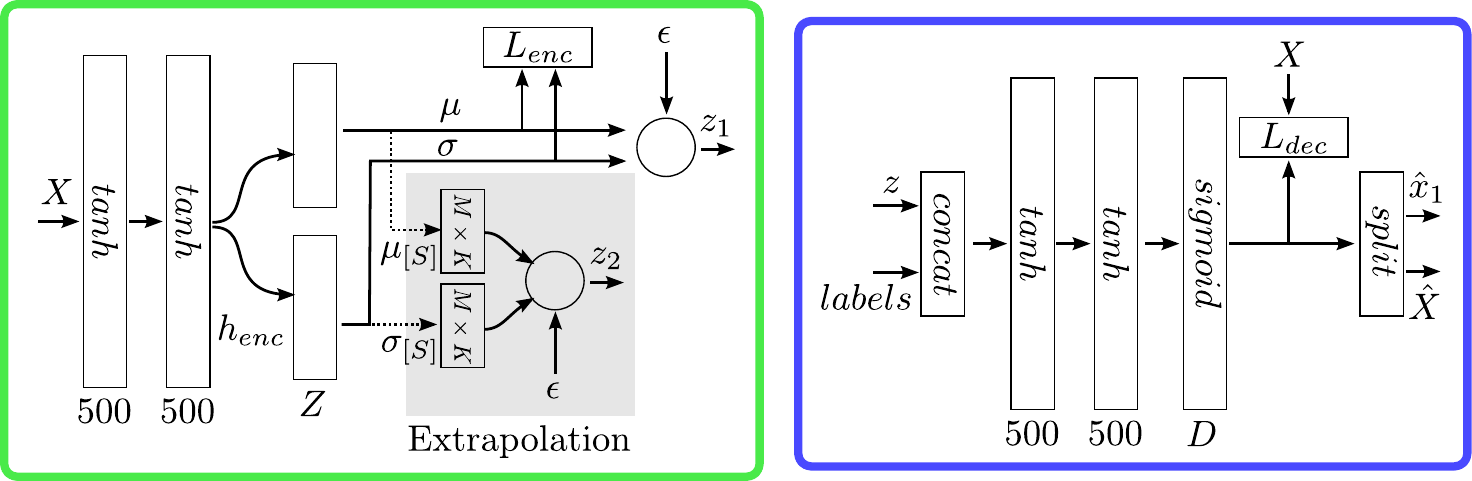}
\end{center}
\vspace{-6pt}
\caption{%
  \textbf{Extrapolation encoder and image generator.} \textbf{Left}: The extrapolation encoder produces two sets of latent distributions. One set ($z_1$) corresponds to a standard VAE encoder. The second set ($z_2$) is produced by our extrapolation layer (shaded portion). See Sections~\ref{sec:variationalencoder},~\ref{sec:extrapolation} for details. \textbf{Right}: The image generator produces images for both sets of codes. All generated images are compared against ground truth. The $z_2$ images are the analogical reasoning output. See Section~\ref{sec:modelgenerator}.
}
\label{fig:detail1}
\label{fig:detail2}
\end{figure}


\subsection{Variational encoder}
\label{sec:variationalencoder}

The encoder takes an image $x$ and produces a latent code $z$
(Figure~\ref{fig:detail1}).  In particular, we describe the encoder as a function
$g$ which maps $x$ to $\mathcal{N}(\mu,\Sigma)$, a multivariate normal distribution. $\Sigma$ is a diagonal covariance matrix --- its diagonal is the vector $\sigma^2$. The latent code is
$z \sim g(x) \in \Reals^Z$.

We implement $g(x)$ as a multilayer perceptron with
two hidden layers of 500 nodes and scaled hyperbolic
tangents~\cite{lecun1989generalization}: $h(a) = 1.7159 \tanh
\frac{2}{3} a$. The activations of the output layer is called the
hidden encoder state, $h_{enc}$. We extract $\mu$ and $\sigma$ from
$h_{enc}$ as $\mu = h_{enc} W_1 + b_1$ and $\log \sigma = \frac{1}{2}
(h_{enc} W_2 + b_2)$ where $W_i$ and $b_i$ are learned weights and
biases.
Then a reparameterization is
$z = \mu + \sigma\epsilon$ ($\sigma$ and $\epsilon$ are multiplied element-wise). $\epsilon \sim
\mathcal{N}(0,I)$ is an auxilliary noise variable.
We take the latent distribution prior
to be an isotropic Gaussian
$\mathcal{N}(0,I)$.
Then the
deviation of a latent distribution from the prior is the KL
divergence, $D_{KL}(\mathcal{N}(\mu,\sigma^2)\;||\;\mathcal{N}(0,I))$.
This is an objective function for our model which regularizes
learned latent distributions.
The encoder loss
is $$L_{enc} = -\frac{1}{2} \sum 1 + 2 \log \sigma - \mu^2 -
\sigma^2.$$
During training, every member of a style
set is encoded as $z_1=g(X) \in \Reals^{M \times Z}$. The preceeding
summarizes how latent codes are produced by a standard VAE and we refer the
reader to~\cite{kingma2013auto} for details.
Our network also produces $z_2$, extrapolated codes from a
subset of $X$.

\subsection{Extrapolation layer}
\label{sec:extrapolation}
The extrapolation layer takes subsets of $\mu$ and $\sigma$ and extrapolates them to cover
every member of our style set. Let $\mu_{[S]} \in \Reals^{K \times Z}$ be the subset of $\mu$ corresponding to $X_{[S]}$ and likewise for $\sigma_{[S]}$.
We introduce a linear mixing layer that takes $K$
latent distributions, mixes them and generates $M$ latent distributions
which are sampled to get latent codes. The mixture is $\mu_{mix} =
W_3 \mu_{[S]}$ and $\sigma_{mix} = W_4 \sigma_{[S]}$ where
$W_3$
and $W_4$ are $M \times K$ matrices. Then $z_2 = \mu_{mix}+\sigma_{mix}\epsilon \in \Reals^{M \times Z}$. The extrapolation layer allows the model to fit a latent distribution to each member of the style set.

\subsection{Image generator}
\label{sec:modelgenerator}

The image generater maps $(z, label)$ to
an image $\hat{x}$ (Figure~\ref{fig:detail2}). $z$ is a latent code and $label$
is a conditional variable (one of $M$ classes, one-hot encoded) which
controls the content of the output. We implement $f$ as a three-layer
multilayer perceptron which takes the concatenation of $z$ and $label$
as the input vector. The hidden layers (500 nodes) have scaled hyperbolic tangents
and the output layer has a sigmoid nonlinearity.
The output layer has a number of nodes equal to $D$,
the product of the image dimensions. We apply $f$ to both $z_1$ and $z_2$
to produce $\hat{x}_1 = f(g(X))$ and $\hat{x}_2 = f(z_2)$. $\hat{x}_1$
corresponds to running the model as an autoencoder so we want $\hat{x}_1$
to be $X$. The analogical reasoning output of the model is
$\hat{X} = \hat{x}_2$ which we also want to be $X$.
Outputs are compared to $X$ with a
structured similarity objective function.

\subsection{Structured similarity objective}
\label{sec:ssim}
We want to use SSIM to evaluate image quality.
However, calculation of SSIM involves costly Gaussian filtering on local neighborhoods. Instead,
we define a global SSIM
objective function, $SSIM^*$.
Its computation depends on global means, $\mu_1$ and $\mu_2$, biased
variances, $\sigma_1$ and $\sigma_2$, and covariance, $\sigma_{12}$. The
similarity of two images, $x_1$ and $x_2$, is calculated as
$$SSIM^*(x_1,x_2) = \frac{(2 \mu_1 \mu_2 + c_1)(2 \sigma_{12} + c_2)}{(\mu_1^2+\mu_2^2+c_1)(\sigma_1^2+\sigma_2^2+c_2)}$$
where $c_1 = 0.01$ and $c_2 = 0.03$.
Our per-channel means and variances are scalars since calculations are
summed over all pixels rather than local neighborhoods.
We refer the reader to~\cite{wang2004image} for explanations of the terms.
The objective is a similarity measure so we use the negative
similarity as the decoder loss function: $$L_{dec} = -\frac{1}{2}(SSIM^*(\hat{x}_1,X) + SSIM^*(\hat{X},X))$$

\subsection{Content classifier}
\label{sec:cls}
We want to be able to alter the content of a generated image so that
it differs from that of an input image. This is easily accomplished
if $z$ codes are invariant to content (i.e., class). In the
case of fonts, this would make the latent space a style embedding.
\cite{ganin2015unsupervised} introduced a novel method for using an
adversarial network to make an embedding space invariant to domain. We
adapt their idea to make our space invariant to content.

The classifier takes a $z$ code and classifies it as one of $M$ classes
with a softmax loss, $L_{cls}$. We implement the classifier as a three-layer scaled
hyperbolic tangent multilayer perceptron with 500-node hidden layers. The
classifier is trained normally, but its gradients are multiplied by
$-\lambda$ when they are propagated back to the green zone. We set
$\lambda = 1$. The intuition behind this gradient reversal is that the
weights and biases of the green zone will move so as to make the classification
task impossible. If this is achieved then the classifier will be unable
to infer content from a $z$ code. A strong indicator that the space is
invariant to content.

We train the content classifier on $z_1$, the codes from the entire style set. The adversary is
trained in a second stage after the encoder and decoder have been trained. Empirically we find that content is easily inferred from $z$ codes
(Section~\ref{sec:experimentadversary}).  Regardless, we find that
including this adversary improves performance on the test set.

\subsection{Imposter detector}
\label{sec:img}
\label{sec:imp}

We also experimented with applying an adversary
to the reconstruction. The intuition is that the imposter detector will
learn discriminating visual features and gradient reversals of those
detections may provide a supplemental reconstuction loss signal.

The imposter detector is the same architecture as the content
classifier. Its two-way softmax loss is $L_{imp}$. We train the imposter
detector on minibatches of $X$ and $\hat{X}$ so that it sees an equal
number of true and imposter samples.

Overfitting is guaranteed since the detector can memorize the true
samples from the training set. Therefore, we test the detector against
the validation set and freeze the purple zone weights and biases when performance
drops on the validation set.

\subsection{Training}
\label{sec:training}
We train the network in two stages. The first stage consists of only the encoder
and decoder (green and blue zones).
The second-stage includes the adversaries. We evaluate validation performance
every 10 epochs and stop when loss on the validation set has not improved
for 100 epochs. We implement our model in Theano~\cite{theano1, theano2}
and optimize with SGD with Adagrad learning updates.

\medskip
\noindent{\bf First stage.} Each minibatch is a complete style set. Therefore,
the encoder produces a $z_1$ code for every member of the style set. 
The extrapolation layer is applied to produce the $z_2$ codes as mixtures
of a subset of the style set. In total $2M$ codes are produced, $M$
codes which correspond directly to each member of the style set, and $M$ codes
which are derived from $X_{[S]}$. All codes are concatenated with appropriate
class labels and passed to the generator which generates $2M$ images. Half of
the images, $\hat{x}_1$, correspond to running the network in autoencoder mode
(the same as a regular VAE). The other half are $\hat{X}$, the analogical
reasoning result. Reconstruction losses are applied to both $\hat{x}_1$ and
$\hat{X}$. The weighted first-stage loss is: $$L_1 = \alpha L_{enc} + \beta L_{dec}$$

\medskip
\noindent{\bf Second stage.} When the full network
is trained its green and blue zones are initialized
with weights and biases from a first-stage trained network. The imposter detector
receives an input of $2M$ images. Half the images are $\hat{X}$ the
other half are $X$. Input images are concatenated with appropriate
class labels. The content classifier is given $M$ $z_1$ codes, the
codes produced by $X$. Training proceeds as normal except after each
evaluation the imposter detector is evaluated against the validation
set, labeled as non-imposters. It is expected that accuracy on the
validation set will rise monotonically. When validation accuracy drops
we freeze the purple layer for the rest of
training. Aside from freezing those weights, training continues as
normal. The weighted second-stage loss is: $$L_2 = \alpha L_{enc} +
\beta L_{dec} + \gamma L_{cls} + \delta L_{imp}$$

\section{Experiments and Results}
\label{sec:experiments}

\begin{table}[tb]
\begin{center}
\tablesize{%

  \begin{tabular}{|l|l|l|l|l|}
  \hline
  \textbf{Model} & \textbf{DSSIM Val} & \textbf{DSSIM Test} \\
  \hline
  \kw~\cite{kingma2014semi} & 0.1149 & 0.1276 \\ 
  Ours-SSIM                   & 0.0915 $\pm$ 0.0005 & 0.1018 $\pm$ 0.0005 \\ 
  Ours-Adv                    & 0.0892 $\pm$ 0.0002 & 0.0990 $\pm$ 0.0002 \\ 
  \hline
  Ours-Adv/prior              & 0.1002 $\pm$ 0.0055 & 0.1131 $\pm$ 0.0050 \\ 
  Ours-L2                     & 0.0978 $\pm$ 0.0002 & 0.1089 $\pm$ 0.0003 \\ 
  Ours-Adv/avg                & 0.0957 $\pm$ 0.0001 & 0.1011 $\pm$ 0.0028 \\ 
  Ours-Adv/cls                & 0.0893 $\pm$ 0.0003 & 0.0994 $\pm$ 0.0003 \\ 
  \hline


  \end{tabular}
}
\end{center}
\vspace{-4pt}
\caption{
  \textbf{Alphanumeric performance results.} \textbf{Top}: We report dissimilarity (mean DSSIM, lower values are better) over 62 alphanumeric classes. Ours-SSIM is our model without the adversaries and Ours-Adv is our full model. See Section~\ref{sec:performance} for details. \textbf{Bottom}: We report performance for variations of our model. Ours-Adv/prior is the full model parameterized to match the prior of \kw, Ours-L2 is our model without adversaries and our $SSIM^*$ loss replaced with L2, Ours-Adv/avg is our full model but without the extrapolation layer and Ours-Adv/cls is our model without the image imposter adversary.}
\label{tab:performance}
\vspace{-6pt}
\end{table}

\begin{table}[htb]
\begin{center}
\tablesize{%

  \begin{tabular}{|l|l|l|l|l|}
  \hline
  \textbf{Model} & \textbf{DSSIM Val} & \textbf{DSSIM Test} \\
  \hline
  \kw~\cite{kingma2014semi} & 0.0883 & 0.0923 \\ 
  Ours-SSIM                   & 0.0711 $\pm$ 0.0002 & 0.0775 $\pm$ 0.0003 \\ 
  Ours-Adv                    & 0.0707 $\pm$ 0.0001 & 0.0771 $\pm$ 0.0000 \\ 
  \hline


  \end{tabular}
}
\end{center}
\vspace{-4pt}
\caption{
  \textbf{Digit performance results.} We report dissimilarity for 10 digit classes (lower is better). See Section~\ref{sec:digits} for details.}
\label{tab:digitperformance}
\vspace{-6pt}
\end{table}

\begin{table}[htb]
\begin{center}
\tablesize{%
  \begin{tabular}{|l|c|c|c|c|c|}
  \hline
  Input letter: & \textbf{H} & \textbf{Y} & \textbf{f} & \textbf{g} & \textbf{s} \\
  \hline
  \textbf{DSSIM Val}  & 0.0881 & 0.0898 & 0.0990 & 0.0931 & 0.0934 \\
  \textbf{DSSIM Test} & 0.0985 & 0.1031 & 0.1081 & 0.1008 & 0.1016 \\
  \hline
  \end{tabular}
}
\end{center}
\vspace{-4pt}
\caption{
  \textbf{Input selection.} We report dissimilarity on alphanumerics with a randomly selected input class for Ours-Adv. Proper selection of the input class does affect performance. See Section~\ref{sec:randomly} for details.}
\label{tab:randomly}
\vspace{-6pt}
\end{table}

\begin{table}[tb]
\begin{center}
\resizebox{0.7\columnwidth}{!}{%
  \begin{tabular}{|l|l|c|c|c|c|c|}
  \hline
  \textbf{Model} & \textbf{Dataset} & \textbf{$Z$} & \textbf{$\alpha$} & \textbf{$\beta$} & \textbf{$\gamma$} & \textbf{$\delta$} \\
  \hline
  \kw~\cite{kingma2014semi}   & Alphanum & 20 & & & & \\
  Ours-L2                     & Alphanum & 50 & 1 & 100             &     &     \\
  Ours-SSIM                   & Alphanum & 50 & 1 & $2 \times 10^5$ &     &     \\
  Ours-Adv/cls                & Alphanum & 50 & 1 & $2 \times 10^5$ & 200 &     \\
  Ours-Adv                    & Alphanum & 50 & 1 & $2 \times 10^5$ & 200 & 200 \\
  Ours-Adv/avg                & Alphanum & 50 & 1 & $2 \times 10^5$ & 200 & 200 \\
  Ours-Adv/prior              & Alphanum & 50 & 1 & 600             & 200 & 200 \\
  \hline
  \kw~\cite{kingma2014semi}   & Digit    & 20 & & & & \\
  Ours-SSIM                   & Digit    & 20 & 1 & 2000 &     &     \\
  Ours-Adv                    & Digit    & 20 & 1 & 2000 & 25 & 25 \\
  \hline
  \end{tabular}
}
\end{center}
\vspace{-4pt}
\caption{
  \textbf{Model parameters.} The parameters of various models used in experiments (Section~\ref{sec:experiments}) are described in Section~\ref{sec:model}.}
\label{tab:params}
\vspace{-6pt}
\end{table}

\begin{figure}[htb]
\begin{center}
  \includegraphics[width=0.95\linewidth]{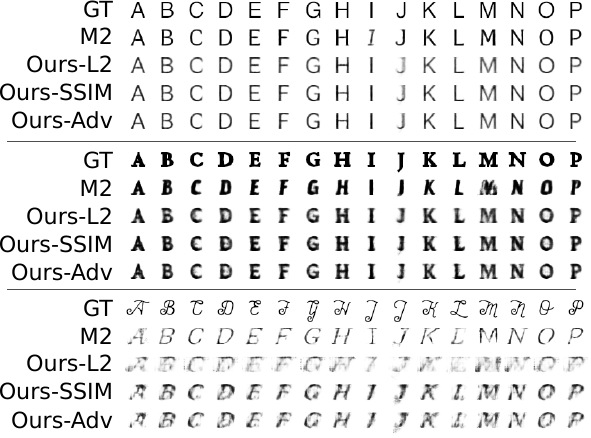}
\end{center}
\vspace{-6pt}
\caption{%
  \textbf{Alphanumeric visual comparisons.} We visualize ground truth (GT), VAE (\kw)
  and three variations of our model (Ours-L2, Ours-SSIM, Ours-Adv).
  The input image is `A'. We want to compare fonts
  that have the same relative quality across models so we ranked the test set
  for each model then chose fonts which had similar rankings.  The first
  font appears in the top-10 of all four models, the second font appears
  in the \nth{6} decile and the third font appears in the bottom-10.
  \kw does not always maintain a consistent slant (e.g., `I' in the first and third fonts) and Ours-L2
  is blurry. Ours-Adv has less ghosting compared to Ours-SSIM (e.g., letters `C' and `D', second font).}
\label{fig:psnrssim}
\end{figure}

\begin{figure}[htb]
\begin{center}
  \includegraphics[width=0.75\linewidth]{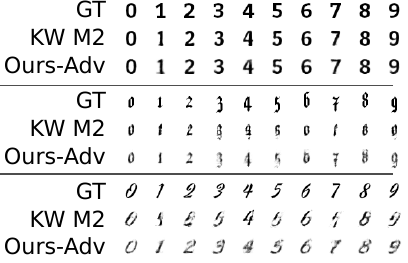}
\end{center}
\vspace{-6pt}
\caption{%
  \textbf{Digit visual comparisons.} We visualize ground truth (GT), VAE (\kw)
  and our full model (Ours-Adv). The input image is `0'. Fonts were selected as in Figure~\ref{fig:psnrssim}.
  The first font appears in the top-10 of all
  four models, the second font appears in the \nth{7} decile and the third
  font appears in the \nth{2} decile. The odd baseline of the second
  font is better captured by our model and our `1' for the third font
  is correctly slanted.}
\label{fig:digits}
\end{figure}

\subsection{Alphanumeric performance}
\label{sec:performance}
To measure performance we report mean DSSIM for the M2 model of~\cite{kingma2014semi}
(\kw), our model without adversaries (Ours-SSIM), and our full model
(Ours-Adv).

Mean DSSIM is the mean structural dissimilarity over the set of all images in the style
set produced by giving the network one image as input. Here, $DSSIM = (1-SSIM)/2.$ where $SSIM$ is
Oliveira's implementation (means computed on Gaussian
filtered neighborhoods) of Wang's SSIM. Note, the reconstruction of $X_{[S]}$ (the input images)
can be trivially perfect by simply echoing back the input rather
than using the network outputs. Since we want to evaluate the
performance of the network we use the network outputs. This gives a more
comprehensive measure of network performance.

Network parameters are given in Table~\ref{tab:params}.  We use a SGD
solver with a learning rate of 0.001, weight decay 0.01, momentum 0.9
and L2 weight regularization. The input is $X_{[S]} = \{\text{`A'}\}$.

We test with auxiliary noise variable $\epsilon=0$.
Each model was trained three times with different
random weight initializations (zero-centered Gaussian, standard deviation
0.01). Biases are initialized to zero. We report the mean and uncertainty
of the mean (i.e., $\text{stddev}/\sqrt{3}$). We take weights from the best of the three
trials to initialize Ours-Adv for second-stage training.

We find that all variations of our model in all experiments we performed
have lower dissimilarity than \kw (see Table~\ref{tab:performance}). It is important to note that
\cite{kingma2014semi} did not design their network to solve single-image analogies but
rather observed that their model could produce such results. We compare
against their model since we qualitatively believe they represent the
best state-of-the-art for single-image font analogies. Explicitly, our best
model, Ours-Adv, has 22.4\% (val) and 22.4\% (test) lower dissimilarity
compared to \kw.

\subsection{Alphanumerics versus digits}
\label{sec:digits}
Our dataset images are high-quality renderings from font definition files
and have more than 10 classes. To control for the number of
classes we subsampled our dataset to contain only the 10 digit classes. In
Figure~\ref{fig:digits} we compare generated digits versus alphanumerics.
We find that having
less classes makes for an easier analogy problem with Ours-Adv
mean DSSIM dropping by 20.7\% (val) and 22.1\% (test) compared to
alphanumerics. Likewise, \kw dissimilarity drops by 23.2\% (val) and
27.7\% (test). Results are reported in Table~\ref{tab:digitperformance}
and network parameters are given in Table~\ref{tab:params}.
The validation and test performance gap between our model
and \kw (20.0\% and 16.5\%, respectively) is narrower with 10 classes
compared to 62 classes.

\subsection{Linear mixing}
\label{sec:noearly}
Do we need our extrapolation layer? If
we remove our layer then $z$ codes for missing classes are
produced by averaging the $z$ codes of $X_{[S]}$ (when $|S|=1$ this
is equivalent to the method of~\cite{kingma2014semi} which holds $z$
fixed to produce analogies). We call this variation Ours-Adv/avg
in Table~\ref{tab:performance}.

We find that adding our extrapolation layer
(Ours-Adv) lowers dissimilarity by 6.79\% (val) and 2.08\% (test) compared
to Ours-Adv/avg.

\subsection{L2 versus SSIM}
\label{sec:psnrssim}
It is known that SSIM is
nonconvex~\cite{channappayya2008design, brunet2012mathematical} ---
is it possible that L2 is a better optimizing objective
even though we evaluate with SSIM? In
Table~\ref{tab:performance} we compare the two objective functions by
replacing our $L_{dec}$ (Ours-SSIM) with an L2 loss (Ours-L2).
We find that dissimilarity is 6.44\% (val) and 6.52\% (test) lower with our global SSIM
objective.

We also qualitatively compare the results of Ours-L2 to
Ours-SSIM in Figure~\ref{fig:psnrssim}. We found pairs for
comparison by selecting fonts from the test set which are closely ranked when test
fonts are sorted by SSIM for each model.
To conserve space we only show 16 alphanumerics.

\subsection{Matching the prior} The encoder is an estimator of the prior
and the prior loss is an indicator of how dissimilar the estimator is to the prior. \kw 
may be disadvantaged compared to our
model since we favor test set performance over matching the prior. Indeed,
our best performing Ours-Adv model has a prior loss of 317.015 (val)
whereas \kw is 21.806 (val) on alphanumerics. We can approximately
control for prior loss by finding loss weights for our model that produce
approximately the same prior loss as \kw. We found loss weights
for our model (Ours-Adv/prior) such that the prior loss is 20.835
(val). Constrained to matching prior, our model has 12.8\% (val) and 11.4\% (test) lower
dissimilarity than \kw.

\subsection{Adversaries}
\label{sec:experimentadversary}
We compare a network trained without any adversarial layers (Ours-SSIM) to
networks with a class adversary only (Ours-Adv/cls, Table~\ref{tab:performance}) and with both adversary layers (Ours-Adv). We found that adding a class adversary
lowers dissimilarity by 2.40\% (val) and 2.36\% (test). Adding both the class
adversary and the imposter adversary lowers dissimilarity by 2.51\% (val) and
2.75\% (test).

How well does the class adversary impose class invariance on the latent
space? The accuracy of the in-network classifier gives us a lower
bound on classification accuracy. Chance is 1.6\% and Ours-Adv has an accuracy of 85.3\%
(val). Although, it does not succeed at removing class,
it does improve test set performance. It is possible that co-adapted
interactions with the extrapolation layer allow class dependence in the
latent space.

\subsection{Input selection}
\label{sec:randomly}
We arbitrarily selected the first image of the style set as the input
image for our experiments (`A' for alphanumerics and `0' for digits). How
does performance vary with different input images? We ran an experiment
on alphanumerics where we varied the input letter of Ours-Adv (`H',
`Y', `f', `g' and `s' were randomly selected) and found that the best
input in that set is `H' and the worst is `f'.  The dissimilarity of
`f' is 12.4\% (val) and 9.75\% (test) higher than that of `H'. Proper
selection of the input class can significantly impact performance.


\section{Discussion and Conclusion}

In this paper, we explored a new supervised VAE design for the problem
of analogical reasoning of style and content on images. Our experiments show that
extrapolating latent distributions outperforms using a fixed latent
code for generating image analogies. We also find that having a large
number of classes (e.g., all letters and digits) makes for a much more
difficult problem compared to a smaller number of classes (e.g., just
digits). We also note that using a class adversary leads to slightly
improved results.

While our method outperforms the state-of-the-art on our analogies
problem, we note that there is significant room for improvement. Our
method performs well on standard typefaces, yet on more stylized fonts
it can result in blurry or otherwise garbled glyphs, or letters that
do not match the style in all respects (e.g., the middle and bottom
parts of Figure~\ref{fig:psnrssim}), even though it outperforms prior work.
This suggests that this remains a very challenging problem, both
because of the stylized nature of the images, and because of the large
number of classes; these properties make our dataset compelling for
future work.  We believe it would be fruitful to explore the role of
both more training data and more sophisticated generative models,
e.g., convolutional neural networks that factor style and content.

\medskip
\noindent{\bf Other domains.} In this work we have focused on the
domain of fonts.  However, our method could be applied to other
domains where images can be factored into content and style (or
similar factors). Examples of these domains include faces (where
images can be decomposed into identity and expression, for instance),
images filtered with different Instagram filters, icons in different
styles (see for instance \url{thenounproject.com}, art, and materials
(e.g., learning to transfer textures from one surface to another). In
the future we plan to explore these additional domains.

\leaveout{
\begin{itemize}

\item The performance discrepancy between low prior and high prior loss
may be an indicator that an isotropic Gaussian is a poor prior for images.

\item For image analogies it is important for test set performance to
extrapolate latent distributions rather than latent codes.

\item Using a dataset with large numbers of classes will help us build
better image analogy models.

\item Promoting class invariance in the latent space with an in-network
adversary is an effective way to improve test set performance.

\end{itemize}
}

{\small
\bibliographystyle{ieee}
\bibliography{main}
}

\newpage
\title{Supplemental}
\author{}
\date{}
\posttitle{\par\end{center}\vspace{-1cm}}
\maketitle

\section{Dataset Preparation}

The dataset source is 1839 TrueType and OpenType font files. We
used FreeType~\cite{freetype} to rasterize the fonts. Each font was
rasterized into a 40 $\times$ 40 pixel box on a baseline which is 3/8 of
the vertical height. Font sizes were chosen such that each letter of the
font fits inside the box (minus a one pixel border) while occupying as
much space as possible. Each letter is horizontally translated so that
its bounding box is centered horizontally.

\section{Visual Comparisons}

The main paper has an abridged figure of alphanumeric comparisons. We
present the full comparisons in Figure~\ref{fig:multi}. This figure also
includes comparisons for multiple inputs (Section~\ref{sec:multi}).

\section{Multiple Input}
\label{sec:multi}

Our model takes any subset $X_{[S]}$ as input. We have limited our
main paper experiments to single inputs so we can compare to previous
work and because it is a more challenging problem. However, it is
interesting to evaluate the effect of multiple inputs (i.e., providing
a few example letters). For example, in the case of fonts we could
require three samples: uppercase, lowercase, and digit
examples. Specifically, we select one input from each subdomain of
alphanumerics so that $X_{[S]} = \{ \text{`A'}, \text{`a'}, \text{`0'}
\}$. In this case, our extrapolation layer produces new latent
distributions as linear mixtures of three latent distributions.

\begin{figure*}[htb]
\begin{center}
  \includegraphics[width=0.95\linewidth]{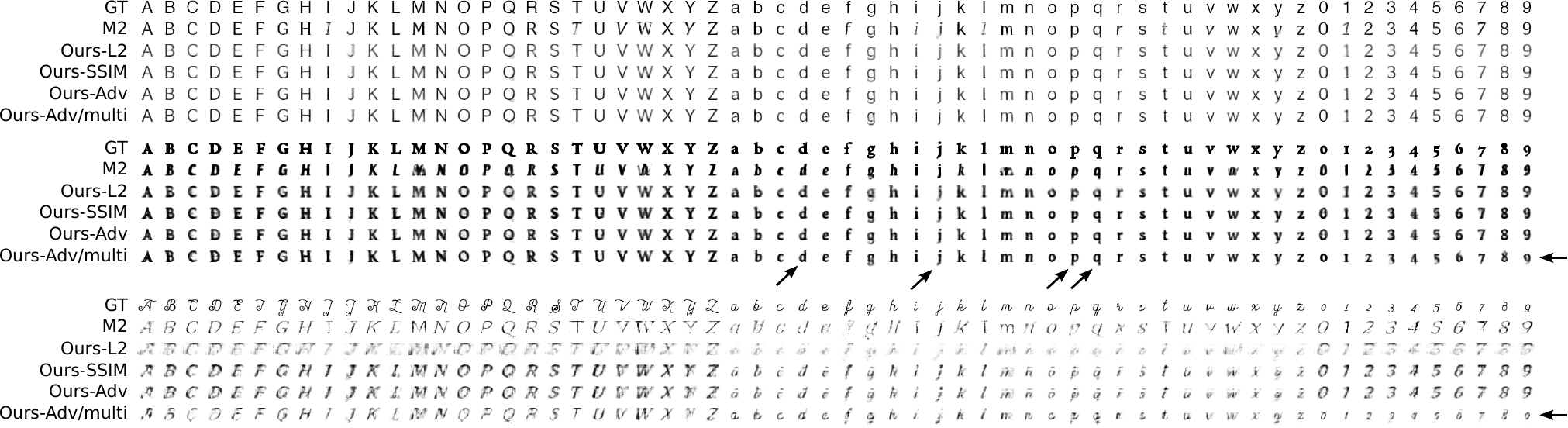}
\end{center}
\vspace{-6pt}
\caption{%
  \textbf{Visual comparisons.} We visualize ground truth (GT),
  single-input VAE (M2), three variations of our single-input model (Ours-L2, Ours-SSIM, Ours-Adv) and
  our multiple-input model (Ours-Adv/multi). The single input image is `A', the additional inputs are `a' and `0'. Test fonts were selected to match Figure 5 of the main paper.
  The first font appears in the top-10 of all
  models, the second font appears in the \nth{6} decile and the third
  font appears in bottom-10. Interesting results for Ours-Adv versus Ours-Adv/multi are marked with black arrows. See Section~\ref{sec:multi} for details.}
\label{fig:multi}
\end{figure*}

The three-input models (Ours-SSIM/multi and Ours-Adv/multi) have the same
training parameters as the single-input models (Ours-SSIM and Ours-Adv).
In Table~\ref{tab:multi} we present our results which show that
multiple inputs improve quality. In Figure~\ref{fig:multi}
we find that with multiple inputs the style of the digits match ground
truth (GT) better (digits indicated with black arrows in the figure) and the lowercase letters of the second font are crisper (e.g., the loop of `d' and the descenders of `j', `p' and `q', as indicated in the figure).

\section{Latent space visualization}

\begin{figure*}[htb]
\begin{center}
  \includegraphics[width=0.95\linewidth]{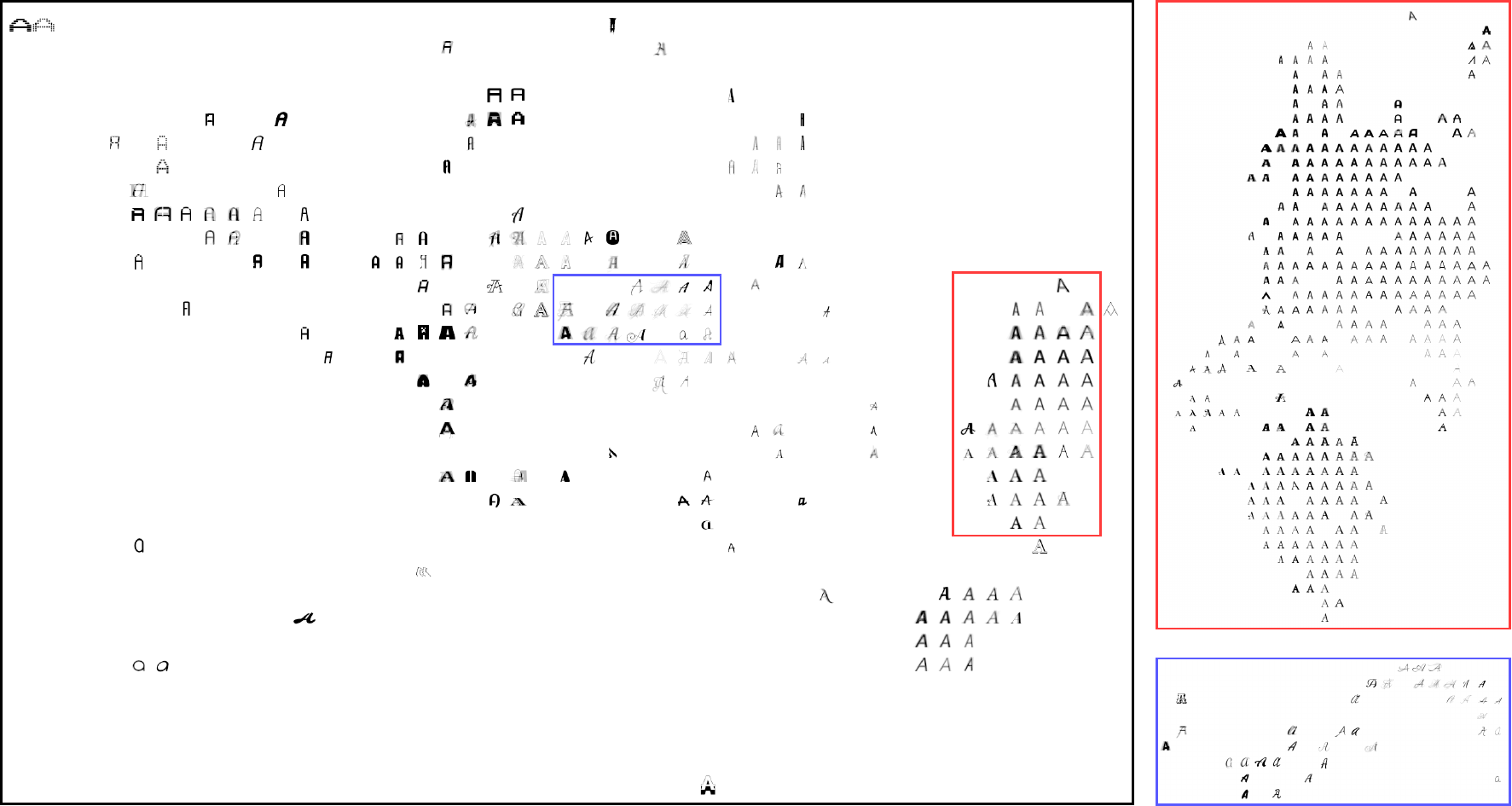}
\end{center}
\vspace{-6pt}
\caption{%
  \textbf{Latent space visualization.} \textbf{Left:} We visualize the latent space with gridded t-SNE. Each grid cell contains the average image of all training samples which are in that cell. Generally, we find that similar styles appear grouped. We examine two regions of interest. \textbf{Top right:} This is a zoom-in of the rightmost region of interest. In this region we find both serif and sans serif fonts in separate clusters. Each cluster is organized by stroke weight (heavier fonts at top). \textbf{Bottom right:} This is a zoom-in of the central region of interest. In this region we find the script fonts. They are also organized by stroke weight (heavier fonts to the left).}
\label{fig:tsne}
\end{figure*}

In Figure~\ref{fig:tsne} we visualize the latent space with
t-SNE~\cite{tsne}. The figure was produced by using t-SNE to project
training samples into a 2-D space. The 2-D space was divided into
non-overlapping cells and each cell is visualized by the average image
of all samples in that cell. Thus, sharp cells contain similar samples
and noisy, blurry cells contain multiple dissimilar
samples. Generally, we find that the space is organized by visual
similarity. This is most clearly seen in the zoomed-in regions of the
figure.

\begin{table}[htb]
\begin{center}
\tablesize{%
  \begin{tabular}{|l|l|l|l|l|}
  \hline
  \textbf{Model} & \textbf{DSSIM Val} & \textbf{DSSIM Test} \\
  \hline
  Ours-SSIM       & 0.0915 $\pm$ 0.0005 & 0.1018 $\pm$ 0.0005 \\ 
  Ours-Adv        & 0.0892 $\pm$ 0.0002 & 0.0990 $\pm$ 0.0002 \\ 
  \hline
  Ours-SSIM/multi & 0.0751 $\pm$ 0.0001 & 0.0854 $\pm$ 0.0002 \\ 
  Ours-Adv/multi  & 0.0735 $\pm$ 0.0001 & 0.0835 $\pm$ 0.0000 \\ 
  \hline
  \end{tabular}
}
\end{center}
\vspace{-4pt}
\caption{
  \textbf{Multiple input alphanumeric results.} Ours-SSIM/multi and Ours-Adv/multi take three letters as input. As expected, dissimilarity is lower with more inputs.}
\label{tab:multi}
\vspace{-6pt}
\end{table}

\end{document}